\documentclass{article} 
\usepackage{iclr2022_conference,times}

\usepackage{amsmath,amsfonts,bm}

\def\eqref#1{equation~\ref{#1}}

\def\1{\bm{1}}

\def\ra{{\textnormal{a}}}

\def\rx{{\textnormal{x}}}

\def\rva{{\mathbf{a}}}

\def\erva{{\textnormal{a}}}

\def\ervx{{\textnormal{x}}}

\def\rmA{{\mathbf{A}}}

\def\vmu{{\bm{\mu}}}
\def\vtheta{{\bm{\theta}}}
\def\va{{\bm{a}}}

\def\ve{{\bm{e}}}

\def\vx{{\bm{x}}}

\def\eva{{a}}

\def\mA{{\bm{A}}}

\def\mH{{\bm{H}}}
\def\mI{{\bm{I}}}
\def\mJ{{\bm{J}}}

\def\mX{{\bm{X}}}

\def\mSigma{{\bm{\Sigma}}}

\DeclareMathAlphabet{\mathsfit}{\encodingdefault}{\sfdefault}{m}{sl}
\SetMathAlphabet{\mathsfit}{bold}{\encodingdefault}{\sfdefault}{bx}{n}
\newcommand{\tens}[1]{\bm{\mathsfit{#1}}}
\def\tA{{\tens{A}}}

\def\tX{{\tens{X}}}

\def\gG{{\mathcal{G}}}

\def\sA{{\mathbb{A}}}
\def\sB{{\mathbb{B}}}

\def\sS{{\mathbb{S}}}

\def\emA{{A}}

\newcommand{\etens}[1]{\mathsfit{#1}}

\def\etA{{\etens{A}}}

\newcommand{\E}{\mathbb{E}}

\newcommand{\R}{\mathbb{R}}

\newcommand{\KL}{D_{\mathrm{KL}}}
\newcommand{\Var}{\mathrm{Var}}

\newcommand{\Cov}{\mathrm{Cov}}

\newcommand{\normltwo}{L^2}
\newcommand{\normlp}{L^p}

\newcommand{\parents}{Pa} 

\usepackage{hyperref}
\usepackage{url}

\usepackage{graphicx}

\title{Surrogate Ensemble Forecasting for Dynamic Climate Impact Models}

\author{Julian Kuehnert, Deborah McGlynn\thanks{Currently at Virginia Tech, Department of Civil and Environmental Engineering, Blacksburg, VA 24061, USA, \texttt{mcglyndf@vt.edu}}, Sekou L. Remy \& Aisha Walcott-Bryant\\
IBM Research Africa\\
Nairobi, Kenya\\
\texttt{julian.kuehnert@ibm.com, \{sekou,awalcott\}@ke.ibm.com} 
\AND
Anne Jones\\
IBM Research Europe\\
Daresbury, United Kingdom\\
\texttt{anne.jones@ibm.com}

}

\iclrfinalcopy 
\begin{document}

\maketitle

\begin{abstract}
As acute climate change impacts weather and climate variability, there is increased demand for robust climate impact model predictions from which forecasts of the impacts can be derived. 
The quality of those predictions are limited by the  climate drivers for the impact models which are nonlinear and highly variable in nature. 
One way to estimate the uncertainty of the model drivers is to assess the distribution of ensembles of climate forecasts. 
To capture the uncertainty in the impact model outputs associated with the distribution of the input climate forecasts, each individual forecast ensemble member has to be propagated through the physical model which can imply high computational costs. 
It is therefore desirable to train a surrogate model which allows predictions of the uncertainties of the output distribution in ensembles of climate drivers, thus reducing resource demands.
This study considers a climate driven disease model, the Liverpool Malaria Model (LMM), which predicts the malaria transmission coefficient R0. 
Seasonal ensembles forecasts of temperature and precipitation with a 6-month horizon are propagated through the model, predicting the distribution of transmission time series.
The input and output data is used to train surrogate models in the form of a Random Forest Quantile Regression (RFQR) model and a Bayesian Long Short-Term Memory (BLSTM) neural network.
Comparing the predictive performance, the RFQR better predicts the time series of the individual ensemble member, while the BLSTM offers a direct way to construct a combined distribution for all ensemble members. 
An important element of the proposed methodology is that accounting for non-normal distributions of climate forecast ensembles can be captured naturally by a Bayesian formulation.
\end{abstract}

\section{Introduction}
\noindent The United Nations' Intergovernmental Panel on Climate Change (IPCC) sixth assessment report indicates that average global temperatures will rise by 1.5$^{\circ}$C above preindustrial levels by 2040 \citep{TheIntergovernmentalPanelonClimateChange2021}. 
This change brings increases in global sea levels, melting of glacial ice, and more extreme precipitation events \citep{TheIntergovernmentalPanelonClimateChange2021}. 
This in turn is expected to bring increases in drought, wild fires, flooding, and variability in the regionality of diseases, and all with increased forecasting uncertainty.

Accurate forecasting of variables dependent on climate drivers remains a challenge. 
The underlying physical processes driving their variability are dynamic and nonlinear, making predictions through mathematical models difficult and resource expensive. 
Machine learning algorithms can help to generate predictions and associated uncertainties without explicitly knowing the underlying processes,  thereby reducing computational costs. 

In this work, we calculate the malaria parasite dynamics from daily temperature and precipitation data based on the Liverpool Malaria Model \citep[LMM,][]{Hoshen2004}. 
The model has been used to understand malaria transmission dynamics, and predict malaria transmission over seasonal and climate change timescales \citep{Jones2010,Caminade2014}. 
Given its non-linear dependence on two climate variables  (temperature and precipitation), this model serves as a useful example with which to explore the process of quantifying uncertainty in an endogenous variable.

Surrogate machine learning models are trained to predict the seasonal variability of the climate driven variables, using the ensembles of temperature and precipitation as inputs or feature variables, and the ensembles of transmission coefficients R$_{0}$ as output or the target variable. 
We implement two types of machine learning algorithms.
For the first type, a Random Forest Quantile Regression (RFQR) model is used. This ensemble learning model was originally developed by \citet{meinshausen2006quantile} and offers a robust, non-linear, and non-parametric way to predict quantile ranges based on training observations. It has been shown to successully calibrate the dynamics of ensemble weather forecasts \citep{taillardat2016calibrated}.
For the second type, a Bayesian Long Short-Term Memory (BLSTM) model is used. Its nature of a recurrent Neural Network (RNN) is well suited for sequential data such as natural language processing and temporal data \citep{Wang2019,Han2020}. 
The LSTM model is a type of RNN that incorporates gated back propagation of data \citep{Wang2020,Han2020} to mitigate the vanishing gradient problem \citep{Khan2018}. 
In our work, the LSTM is implemented as an approximation of a Bayesian NN to capture the uncertainty in the climate forecast ensembles.

\section{Methodology}
\subsection{Climate and malaria transmission data}
While the presented method can be applied to any variable that is highly correlated with climate and weather variability, we apply it to malaria forecasting as an example. 
Climate data was obtained from IBM®\footnote{IBM and the IBM logo are trademarks of International Business Machines Corporation, registered in many jurisdictions worldwide. Other product and service names might be trademarks of IBM or other companies. A current list of IBM trademarks is available on ibm.com/trademark.} PAIRS ECMWF seasonal forecasts \citep{Lu2016,Klein2015,Johnson2019,Crawford2019} for the center of Nairobi, Kenya with coordinates (-1.286389, 36.817222). 
For the 5 year period from 2017 through 2021, 50 ensemble members of daily temperature and precipitation data were obtained in 6 month segments, reflecting the variability of seasonal forecasts. The starting dates of the forecasts are fixed on January 1 and July 1. This ensures that the seasonal peaks of malaria in Nairobi, which occur in April and November, are situated in the middle of the forecast period.
The climate data was then propagated through the Liverpool Malaria Model (LMM) to calculate malaria transmission coefficient R0, using the simplified \textit{steady state} model created by \citet{Jones2007}.

\subsection{Surrogate model description}
The RFQR was introduced by \citet{meinshausen2006quantile}. It extends classical Random Forests, by allowing quantile ranges to be predicted and hence its own prediction uncertainty. 
This is realized by not only keeping the means of the target values in the leaves of the decision trees but storing all the samples of the target values.
Quantiles can then be computed based on these distributions.

The BLSTM was implemented based on the work of  \citet{Zhu2017}. 
First, it combines a classical LSTM with an encoder-decoder in order to extract only representative features in the data, which improves predictive performance on unknown sample patterns. 
Then, following the approach by \citet{gal2016dropout}, Monte Carlo dropout is introduced to randomly remove samples at each layer of the Neural Network. 
This is the process through which the Bayesian NN is approximated, and a posterior distribution of the prediction can be derived by repeating the experiments.

\subsection{Data preparation}
Based on the above described data sources and Malaria Model, the datasets are constructed with following variables: \{Date, Ensemble Member, Temperature [daily mean in degree Celsius], Precipitation (daily mean in mm), Malaria transmission R0\}. Then, the following two training-test datasets are prepared:
\begin{itemize}
    \item \textbf{Dataset 1} comprises 50 ensemble members for a single forecast period of a 6 month horizon from January 1 to June 30, 2021. The training dataset is built from 70\,\% (35) of the ensembles members, while the other 30\,\% (15) of the ensembles members are used for testing.  
    \item \textbf{Dataset 2} comprises 50 ensemble members for eight forecast periods, each of a 6 month horizon starting at either January 1 or July 1 of 2017 through 2021. The dataset is then split into a training part comprising the first 4 years from 2017 through 2020, and a test part comprising the year 2021.
\end{itemize}
\subsection{Model training and testing}
The RFQR is set up with 1000 tree estimators, in which a minimum of 10 samples is required to split an internal node and the minimum number of samples per leaf is allowed to be 1. Loss is measured by the Mean Squared Error (MSE). 
The feature vector is constructed by \{Day of Month, Month, Precipitation, Temperature\} at time step $t_n$, while the target vector contains \{R0\} at time step $t_n$.

The BLSTM is set up as in \citet{Zhu2017}, using the Mean Squared Error (MSE) as the loss function and learning with a rate of 0.01 during 200 epochs. 
For Dataset 1, a batch size of 1024 is used, while for the larger Dataset 2, the batch size is set to 4096. 
Each time series is split into sequences of 50 days while each $50^{th}$ day is predicted based on the previous 49, including the values of previous transmission coefficients R0. 
The feature vector is constructed by \{Day of Month, Month, Precipitation, Temperature, R0\} for time steps $t_i$ to $t_{i+49}$, and the target is \{R0\} at time step $t_{i+50}$, where it is stepping through the length of the whole forecast period. 
200 experiments with Monte Carlo dropout probability of 0.5 are performed to sample the posterior distribution. 

\subsection{Uncertainty estimation}
The uncertainty of the model predictions is quantified by defining a  quantile range from the lower $q_l=15.87$\,\% to the upper $q_u=84.13$\,\%. We have chosen these values because they correspond to minus or plus the standard deviation in the case of a normal distribution.

The quantiles for predicted distribution of each test sample (i.e. for a given date $t_i$ and a given ensemble member $j$) are inferred as follows: \vspace{-0.2cm}
\begin{enumerate}
    \item RFQR: $q_l^j(t_i)$ and $q_u^j(t_i)$ are directly predicted. 
    \item BLSTM: compute $q_l^j(t_i)$ and $q_u^j(t_i)$ from the samples $s_k^j(t_i)$ of the $M$ experiments with random Monte Carlo dropout.
\end{enumerate}
When considering all test ensemble members (i.e. for a given date across all $N$ ensemble members), the combined ensembles are inferred as follows: \vspace{-0.2cm}
\begin{enumerate}
    \item RFQR: compute the combined lower and upper quantile $Q_l$ and $Q_u$ by averaging individual quantiles $Q_l(t_i) = 1/N \sum_j^N q_l^j(t_i)$ and $Q_u(t_i) = 1/N \sum_j^N q_u^j(t_i)$.
    \item BLSTM: derive $Q_l$ and $Q_u$ as for the RFQR. Additionally, we derive a lower and upper quantile range $Qd_l$ and $Qd_u$ directly from the samples $s_k^j(t_i)$ across all experiments $k=1,...,M$ and all ensembles $j=1,...,N$.
\end{enumerate}

\section{Results and discussion}
\subsection{Predictions for individual ensemble members}
\begin{figure*}[h!]
\centering
    \includegraphics[width=0.45\linewidth]{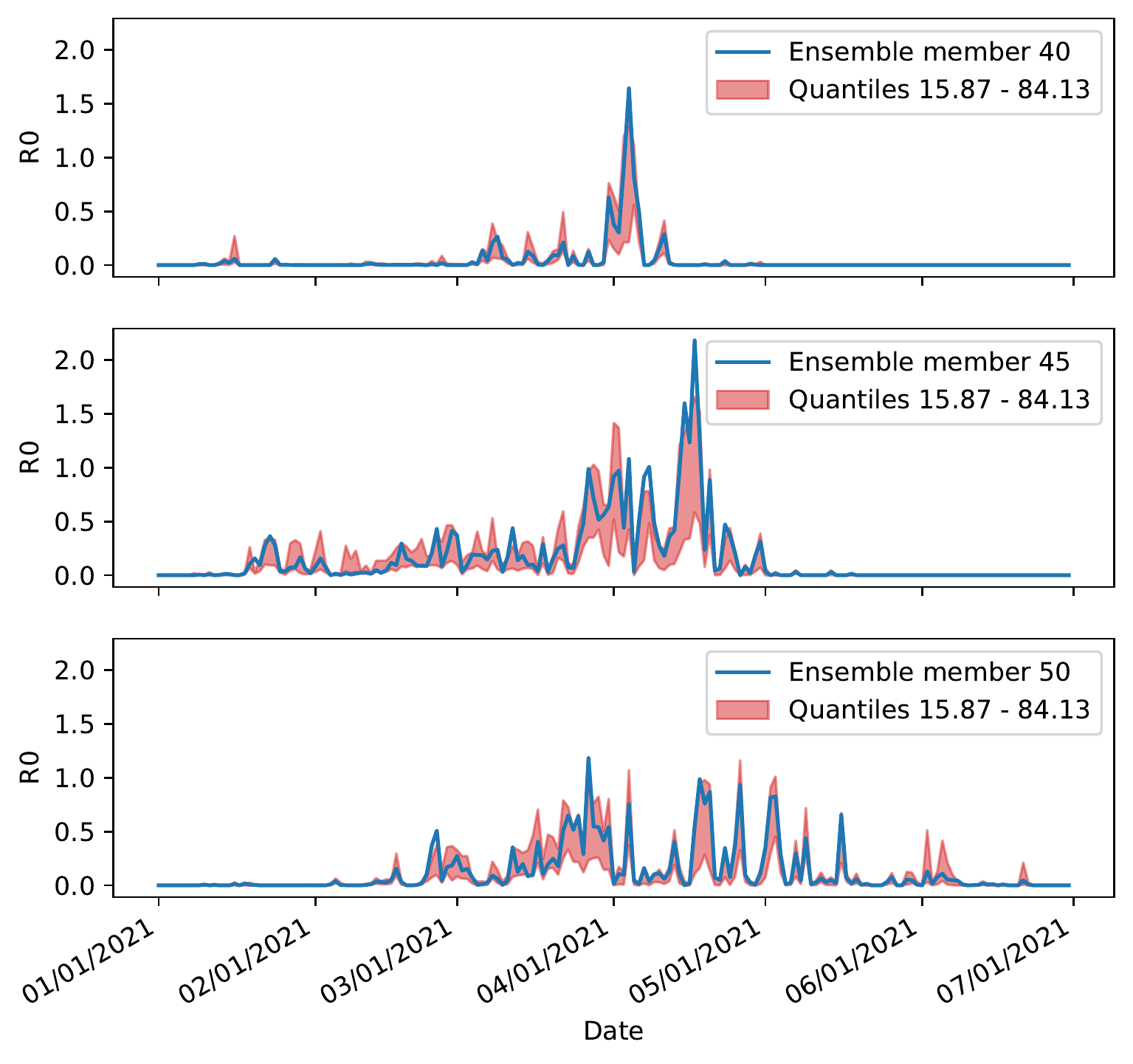}
    \includegraphics[width=0.45\linewidth]{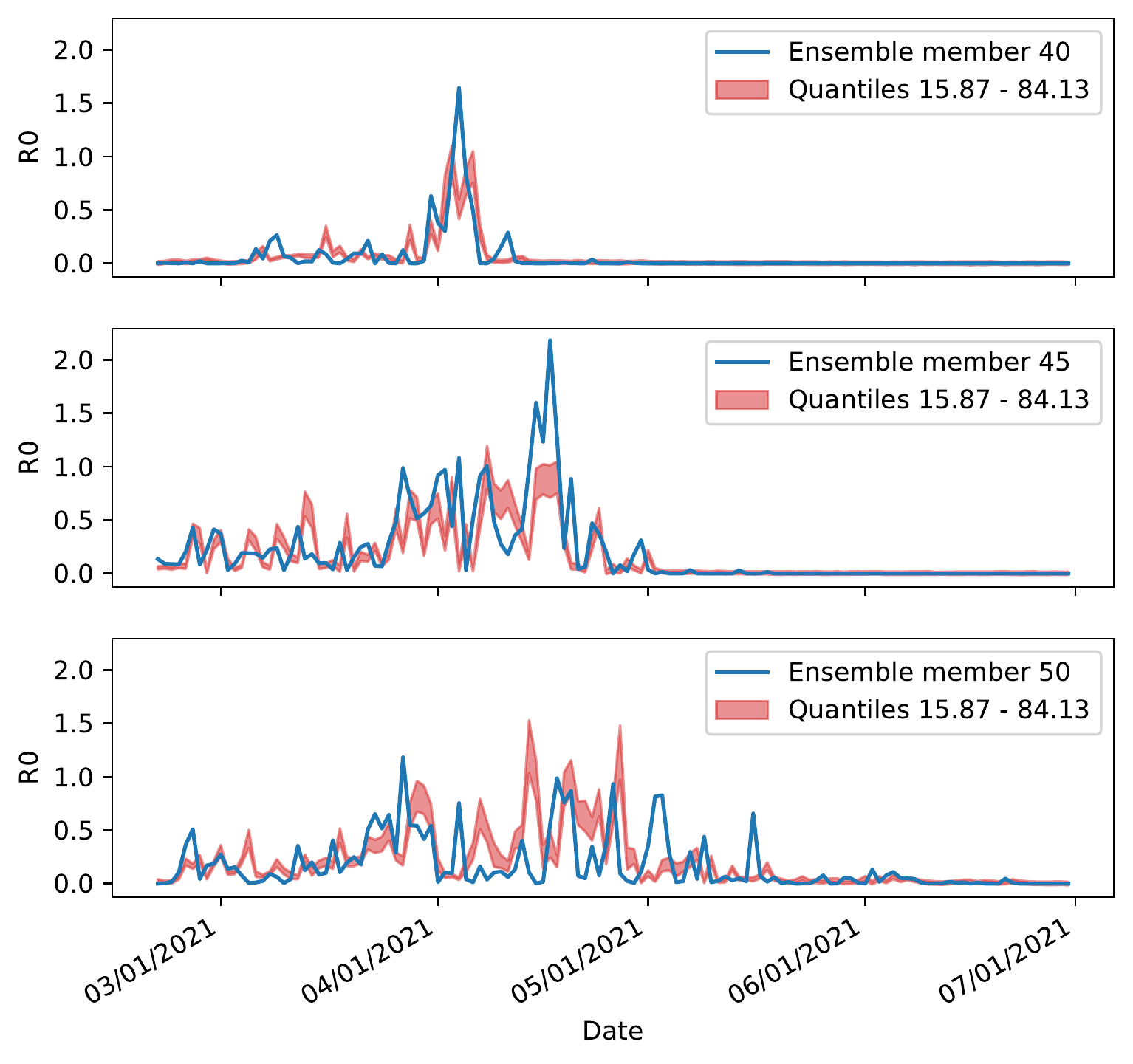}
    \vspace{-0.3cm}
    \caption{Target time series (blue) and predicted quantile range (red) from the RFQR (left) and from the BLSTM (right) for three individual ensemble members.}
    \label{fig:individual_forecasts}
\end{figure*}

For an initial evaluation of the dynamic behavior of the predictions, we randomly pick three individual ensemble members and compare the dynamic quantile range with the true R0 timeseries. 
The comparison is shown in Figure \ref{fig:individual_forecasts} for ensemble members \{40, 45, 50\} and for RFQR (left) and BLSTM (right).
Qualitatively evaluated, the dynamic behavior of the quantile range predicted by the RFQR captures well the time series of the target R0 values. 
Even if the predictions from the BLSTM reflect some of the dynamic characteristics, the predicted quantile range does not succeed in capturing all data points.

\subsection{Predictive performance comparison}
\begin{figure}[h!]
\centering
    \includegraphics[width=0.45\linewidth]{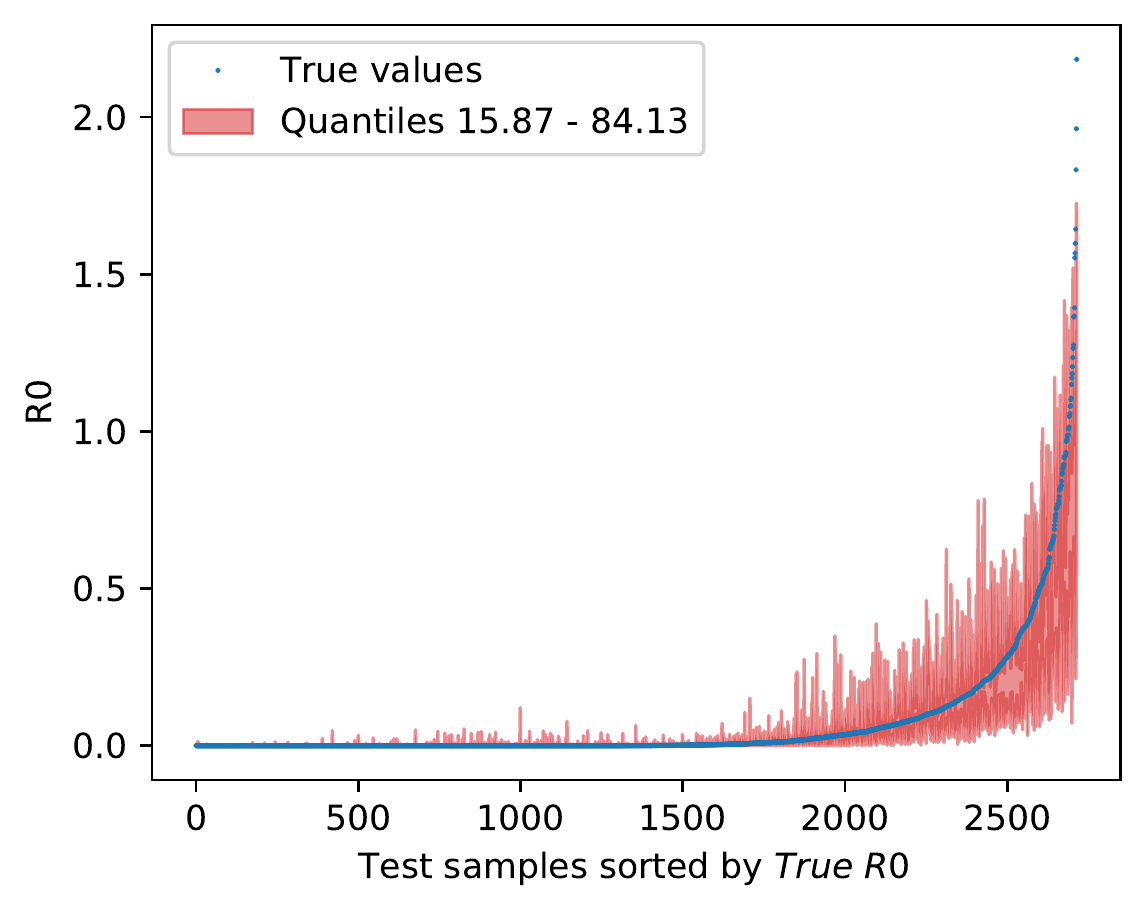}%
    \includegraphics[width=0.45\linewidth]{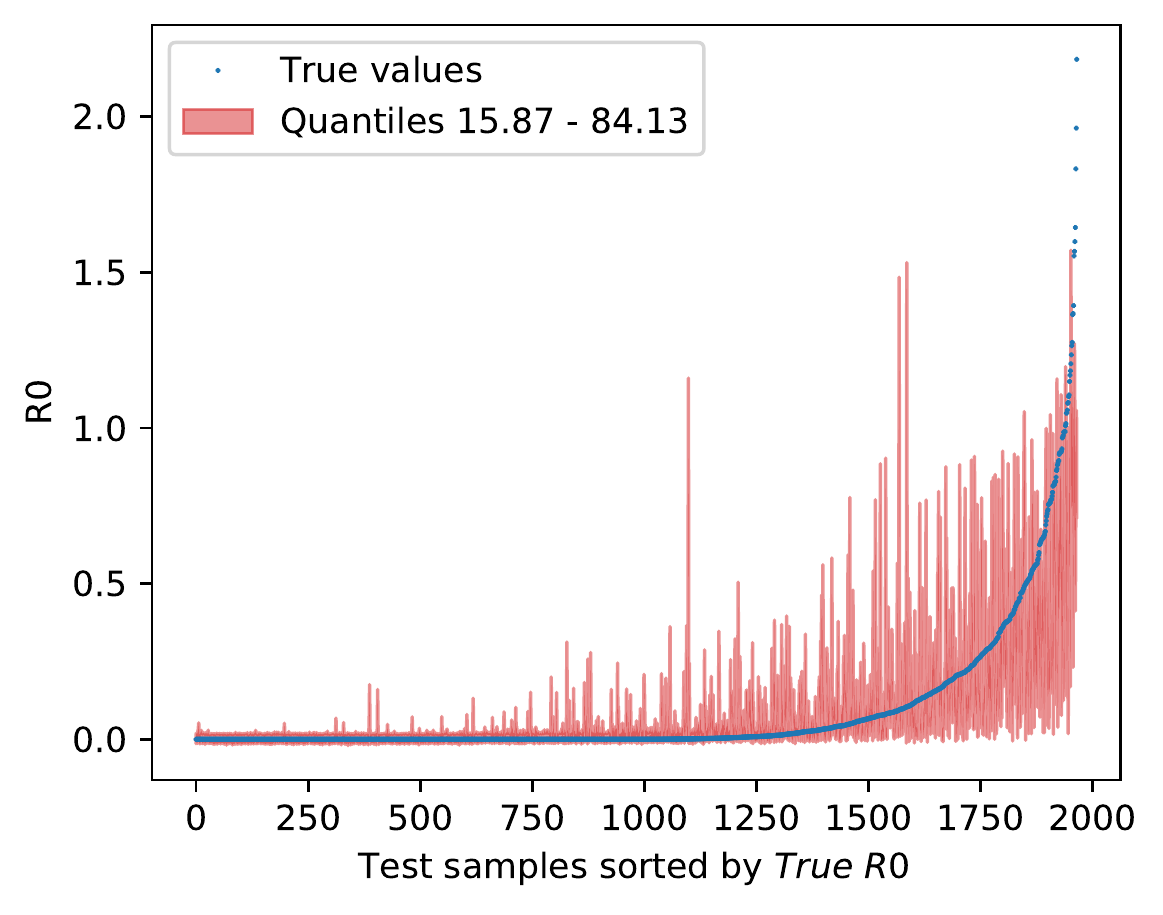}%
    \vspace{-0.3cm}
    \caption{Target values and predicted confidence intervals from RFQR (left) and from BLSTM (right) for all daily samples of the 15 test ensemble members, sorted by the target values.}
    \label{fig:2021_interval_predictions}
\end{figure}

The predictive performance of the two surrogate models is now compared quantitatively for all test ensemble members.
Predictions of the defined quantile range for the 15 test ensemble members in Dataset 1 are shown in Figure \ref{fig:2021_interval_predictions}, using the RFQR (left) and the BLSTM (right). 
The samples are sorted in ascending order of the target values (\textit{True R0}) which are plotted as blue dots.
In both models the prediction uncertainty increases with increasing R0. 
This is possibly based on two factors. 
One, there is a class imbalance in that within the dataset there are more days in the year where the values of R0 are small (smaller than the median value). 
Two, the R0 time series contain larger fluctuations during periods of peak transmissions (i.e. when R0 is high). 
This means that when R0 should be high, there is more variability in the underlying data which possibly increases the prediction uncertainty. 

\begin{figure}[h!]
\centering
    \includegraphics[width=0.45\linewidth]{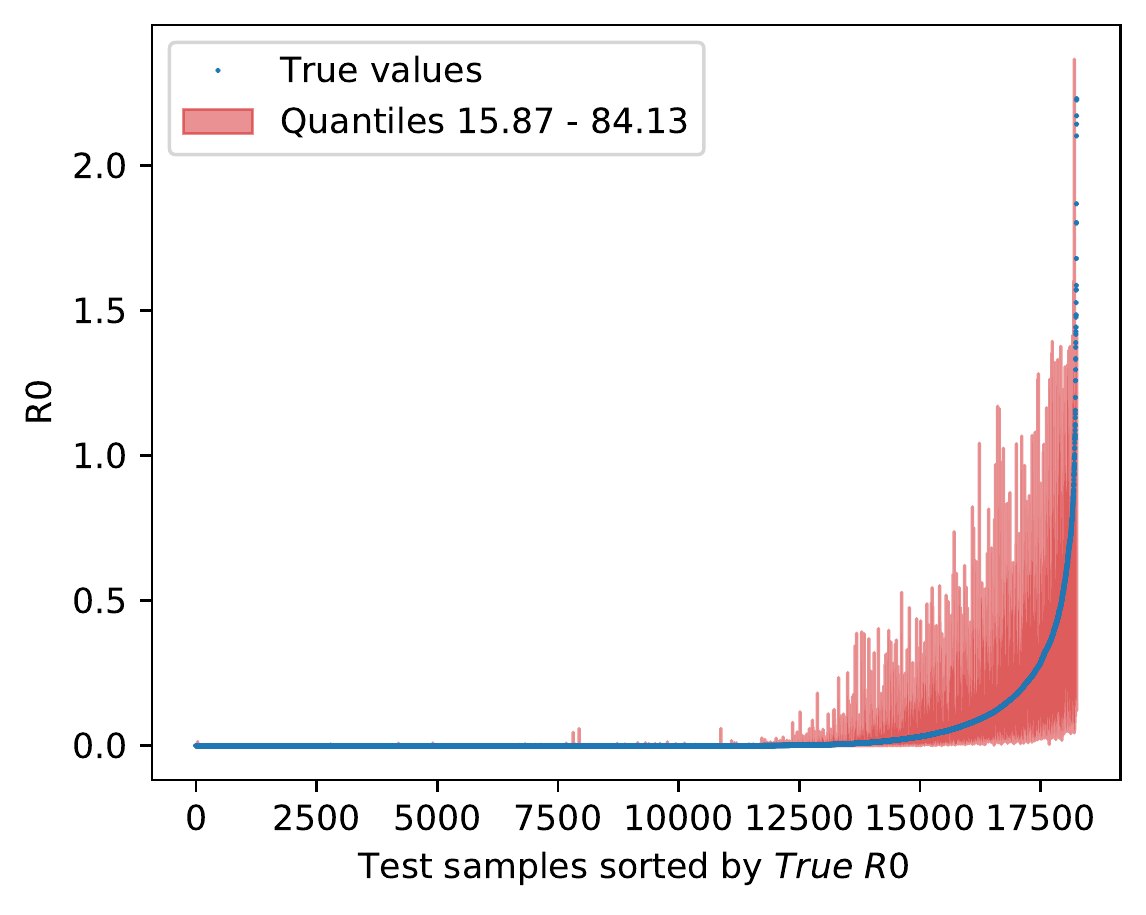}%
    \includegraphics[width=0.45\linewidth]{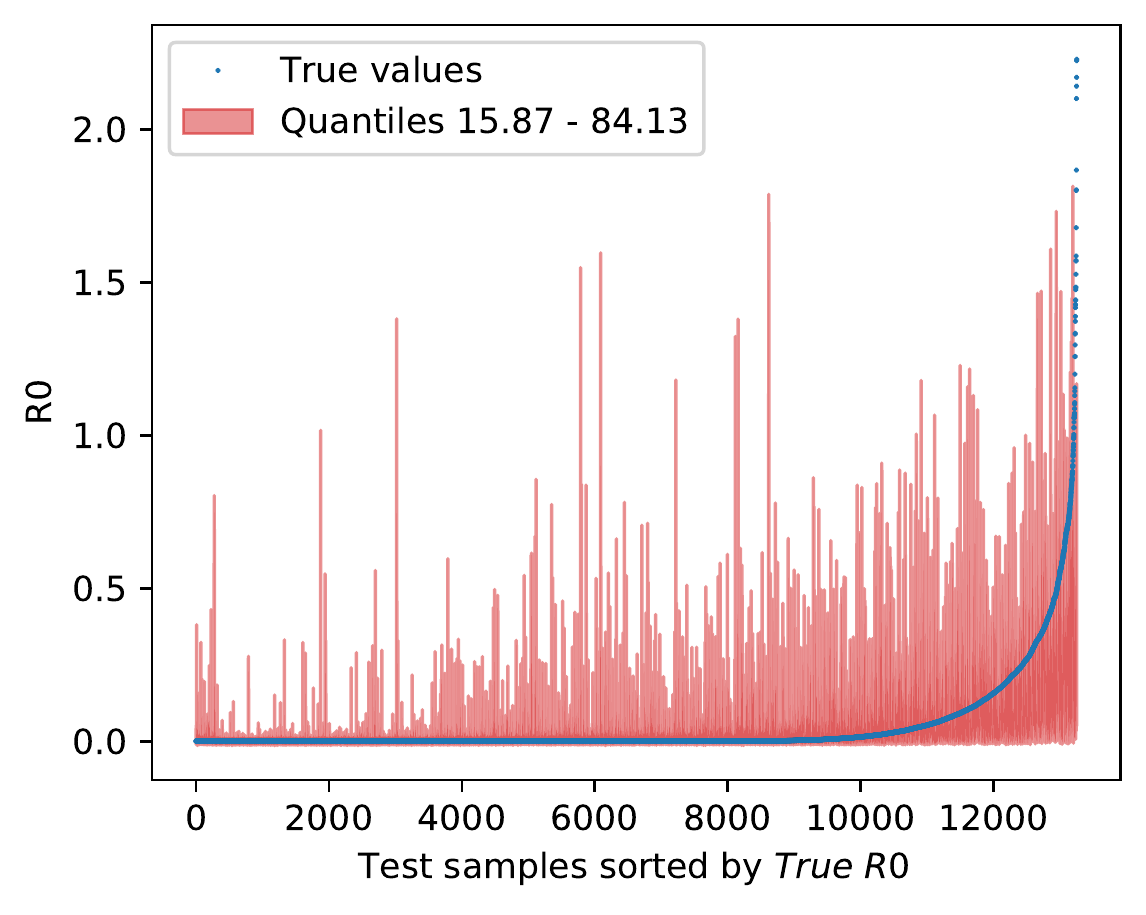}%
    \vspace{-0.3cm}
    \caption{Target values and predicted confidence intervals from RFQR (left) and from BLSTM (right) for all daily samples and all 50 ensemble members of the two test forecast periods in 2021, sorted by the target values.}
    \label{fig:2017-2021_interval_predictions}
\end{figure}

Comparing the two models RFQR and BLSTM, more uncertainty is observed in the BLSTM predictions. 
However, not only is the RFQR quantile range narrower, it also outperforms the BLSTM in capturing the target values in the quantile range. 
This is reflected in Table \ref{tab:hit_rates}, which gives the true data points laying within the quantile range from 15.87\,\% to 84.13\,\% as a percentage of the total number of samples.
\begin{table}[h!]
\centering
\caption{Rates of target values laying within the confidence intervals predicted from RFQR and BLSTM, respectively, for different datasets and differet quantile ranges.}
\label{tab:hit_rates}
\begin{tabular}{lcc}
\hline
               & \textbf{RFQR}     & \textbf{BLSTM }  \\ \hline
Dataset 1, Range $[q_l, q_u]$ & 82.9\,\% & 61.8\,\% \\
Dataset 2, Range $[q_l, q_u]$    & 80.6\,\% & 61.1\,\% \\
Dataset 1, Range $[Q_l, Q_u]$     &     15.5\,\%     &     38.9\,\%     \\
Dataset 2, Range $[Q_l, Q_u]$     &      11.8\,\%    &     34.5\,\%     \\ 
Dataset 1, Range $[Qd_l, Qd_u]$     &    na      &     69.1\,\%    \\ 
Dataset 2, Range $[Qd_l, Qd_u]$  &    na      &      73.9\,\%    \\ \hline
\end{tabular}%
\end{table}\\
Similar results can be seen from the model predictions with Dataset 2. 
Figure \ref{fig:2017-2021_interval_predictions} shows the  predicted quantile range for the test ensemble in year 2021 using the RFQR (left) and BLSTM (right), while the target R0 values are plotted as blue dots. 
Again, the uncertainty of the RFQR is generally lower compared to the BLSTM, while a high percentage of the target values are in the quantile range of the RFQR, outperforming the BLSTM (see Table \ref{tab:hit_rates}). 

\subsection{Ensemble prediction uncertainty}
We now assemble the individual model predictions across all ensemble members on a daily basis. 
First we will calculate $[Q_l, Q_u]$ based on the individual member quantile ranges $[q_l, q_u]$. 
To assess how well the combined quantile range $[Q_l, Q_u]$ captures the variation of the target values R0, the percentage of samples within the range is calculated. 
The corresponding values are listed in Table \ref{tab:hit_rates} for both Dataset 1 and Dataset 2. 
It can be observed that upon averaging the lower and upper quantiles, the combined range is only able to capture a subset of the previously captured samples. 
The process of averaging hence works like a low-pass filter, removing the fluctuations which are necessary to keep the target values within the range. 
Of note, the BLSTM is now able to capture more data points. 
This is because the underlying uncertainty had more variations.

To capture more of the target values, we now calculate the ensemble quantile range $[Qd_l, Qd_u]$ directly from the samples of the model instead of from the previously inferred quantiles. 
In the case of the BLSTM, these samples are generated during repeated predicting with Monte Carlo dropout. 
Analysing the hit rates, see Table \ref{tab:hit_rates}, the percentages of captured data points are again now comparable with the percentages from analysing the individual member samples. 
For visual inspection, figures of the resulting dynamic range are included in Appendix \ref{app:Figures}.

\section{Conclusion and future work}
In this work, we use a Random Forest Quantile Regression (RFQR) model and a Bayesian Long Short-Term Memory (BLSTM) neural network as surrogates for a dynamic climate impact model to predict dynamic quantile ranges from climate ensemble forecasts.
In the application with the Liverpool Malaria Model (LMM) for the prediction of transmission coefficient R0, higher uncertainty was predicted on high R0 values. This is attributed to both a class imbalance in R0 in the dataset and increased fluctuation in larger values of R0.
Even though the BLSTM is designed to predict time series data, the performance of  RFQR is better, likely because it can model the dynamics of the underlying physical processes. 
In order to derive a combined ensemble quantile range, it is observed that averaging the predicted quantile ranges of the individual ensemble members acts as a strong low-pass filter because it only captures a small portion of the target values. 
Computing the ensemble quantile range directly from the underlying predictions samples, instead of the predicted quantiles, allows a high percentage of the target value variations to be captured. 
Nonetheless, other methods may need to be considered to capture the extremes of the distribution, such as for example Extreme Value Theory or Autoregressive Conditional Heteroscedasticity. 

\bibliography{iclr2022_conference}
\bibliographystyle{iclr2022_conference}

\vfill
\pagebreak

\appendix
\section{Appendix: Ensemble prediction uncertainty}\label{app:Figures}
\begin{figure}[ht!]
    \centering
    \includegraphics[width=0.55\textwidth]{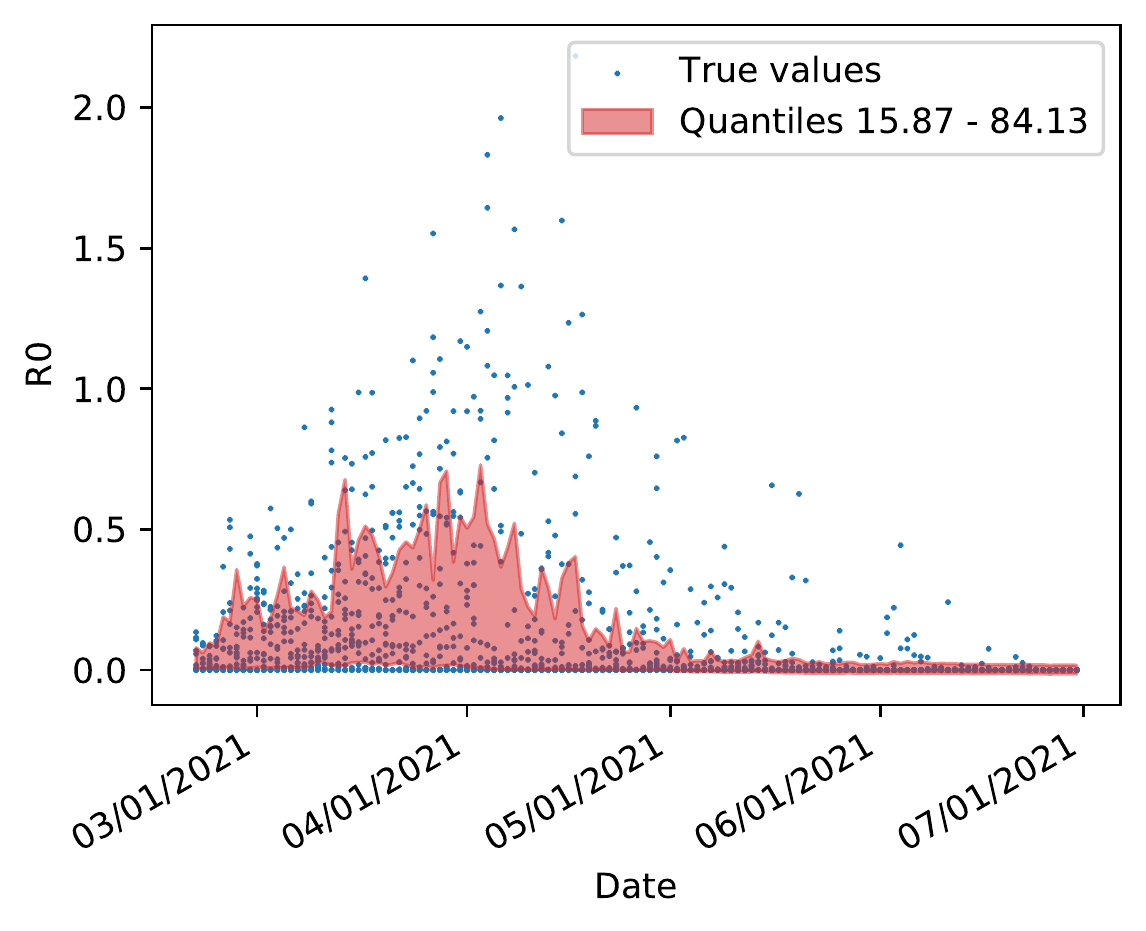}
    \vspace{-0.3cm}
    \caption{Uncertainty prediction using the BLSTM for the 15 test ensembles members for the first half of 2021 in Dataset 1. The model was trained on the other 35 ensemble members for the same forecast period from January 1 to July 1.}
    \label{fig:blstm_2021_agg_samples}
\end{figure}

\begin{figure*}[h!]
    \centering
    \includegraphics[width=\textwidth]{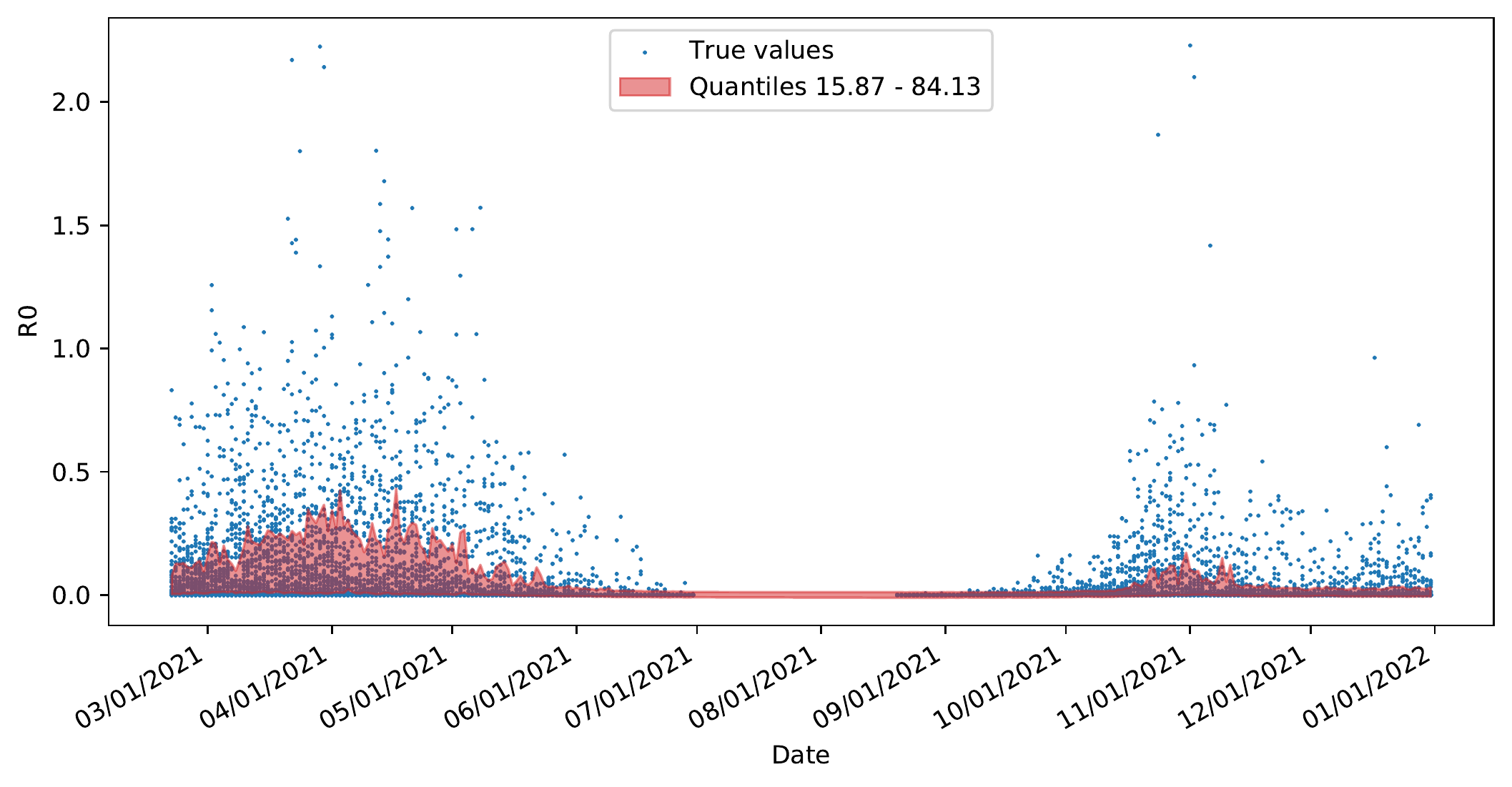}
    \vspace{-0.6cm}
    \caption{Uncertainty prediction using the BLSTM for all 50 test ensemble members in both forecast periods of 2021 in Dataset 2. The model was trained on all 50 ensemble members for 8 six-month forecast periods from 2017 to 2020.}
    \label{fig:blstm_2017-2021_agg_samples}
\end{figure*}

\end{document}